# AI for the prediction of early stages of Alzheimer's disease from neuroimaging biomarkers – A narrative review of a growing field

Thorsten Rudroff[1,2], Oona Rainio[3], and Riku Klén[3]

[1]Department of Health and Human Physiology, University of Iowa, Iowa City, IA, 52242, USA

[2]Department of Neurology, University of Iowa Hospitals and Clinics, Iowa City, IA, 52242, USA

[3]Turku PET Centre, University of Turku and Turku University Hospital, Turku, Finland

**Abstract**

**Objectives** The objectives of this narrative review are to summarize the current state of AI applications in neuroimaging for early Alzheimer's disease (AD) prediction and to highlight the potential of AI techniques in improving early AD diagnosis, prognosis, and management.

**Methods** We conducted a narrative review of studies using AI techniques applied to neuroimaging data for early AD prediction. We examined single-modality studies using structural MRI and PET imaging, as well as multi-modality studies integrating multiple neuroimaging techniques and biomarkers. Furthermore, they reviewed longitudinal studies that model AD progression and identify individuals at risk of rapid decline.

**Results** Single-modality studies using structural MRI and PET imaging have demonstrated high accuracy in classifying AD and predicting progression from mild cognitive impairment (MCI) to AD. Multi-modality studies, integrating multiple neuroimaging techniques and biomarkers, have shown improved performance and robustness compared to single-modality approaches. Longitudinal studies have highlighted the value of AI in modeling AD progression and identifying individuals at risk of rapid decline. However, challenges remain in data standardization, model interpretability, generalizability, clinical integration, and ethical considerations.

**Conclusion** AI techniques applied to neuroimaging data have the potential to improve early AD diagnosis, prognosis, and management. Addressing challenges related to data standardization, model interpretability, generalizability, clinical integration, and ethical considerations is crucial for realizing the full potential of AI in AD research and clinical practice. Collaborative efforts among researchers, clinicians, and regulatory agencies are needed to develop reliable, robust, and ethical AI tools that can benefit AD patients and society.

# Introduction

Alzheimer's disease (AD) is a progressive neurodegenerative disorder characterized by cognitive decline, memory loss, and impaired daily functioning [1]. AD is the most common form of dementia and poses a significant public health challenge, with a rapidly increasing prevalence due to an aging global population [2]. Early diagnosis of AD is crucial for timely intervention, patient management, and the development of disease-modifying therapies [3]. Neuroimaging biomarkers play a critical role in the early detection and monitoring of AD. Structural magnetic resonance imaging (MRI) can assess brain atrophy, functional MRI (fMRI) can reveal alterations in brain connectivity and activation, and positron emission tomography (PET) imaging can provide information on brain metabolism, amyloid deposition, and tau pathology [4]. These neuroimaging techniques have been incorporated into the diagnostic criteria for AD and are widely used in research and clinical settings [5]. Artificial Intelligence (AI) has emerged as a powerful tool for analyzing complex neuroimaging data to aid in the early prediction of AD [6, 7]. AI techniques, such as machine learning and deep learning, can learn patterns and features from large datasets to distinguish between healthy individuals, those with mild cognitive impairment (MCI), and those with AD [8, 9]. These techniques have shown promising results in improving the accuracy, efficiency, and reliability of AD diagnosis and prognosis [10].

This narrative review provides an overview of the current state of AI techniques applied to neuroimaging data for the early prediction of AD. We conducted a targeted literature search, focusing on key subtopics such as single-modality studies using structural MRI and PET imaging, multi-modality studies integrating multiple neuroimaging techniques and biomarkers, and longitudinal studies modeling AD progression. The authors selected relevant studies based on their expertise and familiarity with the literature, aiming to highlight representative works and important trends in the field. While not a systematic review, this article offers a comprehensive perspective on the potential of AI in AD research and identifies current challenges and future directions in this rapidly evolving area.

# Neuroimaging biomarkers in Alzheimer's disease

## Magnetic resonance imaging (MRI)

Structural MRI is widely used to assess brain atrophy in AD. Hippocampal atrophy is one of the earliest and most consistent findings, correlating with memory impairment [11, 12]. Cortical thickness measurements have revealed widespread thinning in AD, particularly in the temporal, parietal, and frontal lobes [13, 14].

fMRI measures brain activity by detecting changes in blood oxygenation. Resting-state fMRI has revealed altered functional connectivity in AD, particularly decreased connectivity in the default mode network (DMN) [15, 16]. Task-based fMRI studies have demonstrated reduced activation in brain regions involved in memory and executive function in AD patients compared to healthy controls [17, 18].

Episodic memory tasks, such as the face-name association task and verbal memory tasks involving learning and recalling word lists, have consistently shown reduced activation in the hippocampus and related medial temporal lobe structures in AD patients [19, 20]. Executive function tasks, like the n-back task assessing working memory and the Stroop task measuring cognitive control and inhibition, have revealed reduced activation in prefrontal and parietal regions in AD, reflecting executive function impairments [21].

*Positron emission tomography (PET)*

Fluorodeoxyglucose (FDG) PET measures brain glucose metabolism, showing a characteristic pattern of hypometabolism in AD in the temporo-parietal cortex, posterior cingulate cortex, and precuneus [22, 23]. FDG-PET has been included in diagnostic criteria for AD [5]. Amyloid PET has revealed that the deposition of amyloid-beta (Aβ) plaques in the brain begins years before symptom onset and this method is therefore a valuable tool for early AD detection [24, 25]. Tau PET has shown that tau pathology spreads from the entorhinal cortex to neocortex, correlating with cognitive decline [26, 27]. Several tracers have been developed for both amyloid and tau PET [28, 29].

## Artificial intelligence techniques for AD prediction

*Machine learning (ML)*

ML approaches can be broadly categorized into supervised and unsupervised methods. Supervised learning algorithms, such as Support Vector Machines (SVM), Random Forests (RF), Logistic Regression (LR), and neural networks learn from labeled training data to make predictions on new, unseen data. These methods have been successfully applied to various neuroimaging modalities, including structural MRI, fMRI, and PET imaging, to classify different stages of AD [9, 30]. Unsupervised learning algorithms, such as clustering and dimensionality reduction techniques, can identify patterns and structure in unlabeled data. These methods, including k-means clustering, hierarchical clustering, and Principal Component Analysis (PCA), have been used to identify subtypes of AD or MCI and to extract latent features from neuroimaging data [31, 32].

Ensemble learning (EL) is a machine learning approach that combines multiple learning algorithms or models to improve overall prediction performance and robustness. In EL, a set of base learners (classifiers or predictors) are trained on the same or different training datasets, and their individual predictions are then aggregated through methods like averaging, voting, or weighted combination to obtain the final prediction. The key idea behind EL is that by utilizing multiple models, the ensemble can achieve better generalization performance and reduce the risk of overfitting. EL approaches can be broadly categorized into homogeneous (using the same learning algorithm) or heterogeneous (using different learning algorithms) methods, and popular techniques include Bagging, Boosting, and RF [33].

*Deep learning (DL)*

DL approaches, based on artificial neural networks, have shown remarkable performance in analyzing complex, high-dimensional data. Convolutional Neural Networks (CNNs) are used to analyze image data, including structural MRI and PET scans, by automatically learning hierarchical features from their input. CNNs have achieved high accuracy in classifying AD, MCI, and healthy controls [34, 35]. Gradient-weighted Class Activation Mapping (Grad-CAM) is a popular explainable AI technique for generating visual explanations from deep learning models [36]. Grad-CAM is a generalization of Class Activation Mapping (CAM) that can work with any type of CNN to produce post hoc local explanations, whereas CAM specifically requires global average pooling. Grad-CAM has been widely used in medical image analysis. For example, Ji [37] used Grad-CAM to show which areas of histology lymph node sections a classifier used to determine the presence of metastatic tissue. [38] applied Grad-CAM to localize small bowel enteropathies on histology images. Windisch et al. [39] employed GradCam to highlight regions of brain MRI that led a classifier to detect the presence of a tumor. Other backpropagation-based approaches for generating saliency maps have also been used in medical image analysis. Böhle et al. [40] compared saliency maps from guided backpropagation [41] and layer-wise relevance propagation (LRP) for identifying Alzheimer's disease regions in brain MRI. They found LRP maps were more specific in highlighting known Alzheimer's regions.

Recurrent Neural Networks (RNNs), designed to handle sequential data, have been applied to model the progression of AD using longitudinal neuroimaging data. RNNs can capture temporal dependencies and predict future cognitive decline or brain atrophy based on past measurements [42, 43]. Autoencoders, a type of unsupervised DL model, have been used for dimensionality reduction and feature extraction from neuroimaging data. By learning a compact representation of the input data, autoencoders can help identify patterns and structure associated with different stages of AD [44, 45].

### Hybrid approaches

Some studies have combined traditional ML and DL techniques to leverage their complementary strengths. For example, DL models can be used to extract high-level features from raw neuroimaging data, which can then be fed into ML classifiers for final prediction [46, 47]. Alternatively, DL models can be used for unsupervised feature learning, followed by ML techniques for clustering or classification [48, 49]. These methods can automatically learn patterns and features associated with different stages of AD, leading to improved diagnostic accuracy and the potential for early intervention and personalized medicine approaches.

For example, Suk et al. [44] proposed a hybrid approach for AD classification using MRI and PET data. They first used a deep auto-encoder to learn high-level features from the neuroimaging data and then employed an SVM classifier to make the final diagnosis. This hybrid approach achieved better classification performance compared to using either method alone.

Another example of a hybrid approach is the work by Liu et al. [46], where they combined a deep learning model (stacked auto-encoders) for feature extraction and an SVM for classification. They applied this method to multi-modal neuroimaging data (MRI and PET) from the ADNI dataset and achieved improved classification accuracy for AD, MCI, and healthy controls compared to single-modality approaches.

Ju et al. [45] proposed a hybrid approach that integrates a CNN for feature extraction and a RNN for modeling temporal dependencies in longitudinal neuroimaging data. By applying this method to MRI data from the ADNI dataset, they demonstrated improved prediction of MCI-to-AD conversion compared to using only CNN or RNN alone.

These examples demonstrate how hybrid approaches can combine the strengths of different ML and DL techniques to improve AD classification and prediction. By leveraging deep learning for feature extraction and traditional ML for classification, or by combining different DL architectures to capture complementary information, hybrid methods have the potential to enhance our understanding of AD and improve diagnostic and prognostic accuracy.

Table 1 summarizes the key AI approaches, techniques, and their applications in AD research, as discussed in the article. It also highlights the neuroimaging data types used and study design and validation methods. This table can serve as a quick reference to grasp the main concepts and components of AI-based AD research covered in this narrative review article.

**Table 1** Overview of AI applications in early AD prediction using neuroimaging data

| AI Approach | Techniques | Applications in AD Research |
|---|---|---|
| Machine Learning (ML) | - Supervised Learning (e.g., SVM, Random Forests, Logistic Regression)—Unsupervised Learning (e.g., Clustering, Dimensionality Reduction)—Ensemble Learning (e.g., Bagging, Boosting, Random Forests) | - AD Classification—MCI-to-AD Conversion Prediction—AD Subtype Identification—Feature Selection and Extraction |
| Deep Learning (DL) | - Convolutional Neural Networks (CNNs)—Recurrent Neural Networks (RNNs)—Autoencoders | - AD Classification—MCI-to-AD Conversion Prediction—AD Progression Modeling—Feature Learning and Extraction |
| Hybrid Approaches | - DL Feature Extraction + ML Classification—DL Unsupervised Learning + ML Clustering/Classification | - Improved AD Classification—Enhanced AD Subtype Identification—Multi-modal Data Integration |
| Neuroimaging Data Types | - Structural MRI—Functional MRI—PET Imaging | - Single-modality Studies: Utilizing one neuroimaging data type for AD prediction—Multi-modality Studies: Integrating multiple neuroimaging data types and other biomarkers for improved AD prediction |
| Study Design and Validation | - Cross-validation (e.g., k-fold, leave-one-out)—Hold-out Validation—External Validation | - Assessing model performance and generalizability—Comparing different AI approaches—Evaluating the robustness of AI models across datasets and populations |

## Applications of AI in early AD prediction

AI techniques have been applied to various neuroimaging modalities, including structural MRI, fMRI, and PET imaging, for early AD prediction. These studies can be broadly categorized into single-modality and multimodality approaches, each with their own strengths and limitations.

### *Single-modality studies: strengths, limitations, and key findings*

Single-modality studies have primarily focused on structural MRI and PET imaging for AI-based AD prediction. Structural MRI-based studies have leveraged AI techniques to extract features and patterns of brain atrophy associated with AD. For example, Basaia et al. [34] achieved high accuracy of 98% in classifying structural MRI data between AD patients and healthy controls with a DL approach. However, the study had a relatively small sample size (295 subjects) and did not include mild cognitive impairment (MCI) individuals, limiting its generalizability. Pan et al. [50] proposes a novel approach combining CNN and EL to identify AD and MCI using structural MRI data from 671 subjects. Their CNN-EL model achieved accuracies of 84% for AD vs. healthy controls (HC), 79% for MCI converters (MCIc) vs. HC, and 62% for MCIc vs. MCI non-converters, outperforming other methods like PCA +SVM and 3D-SENet, while also identifying the most discriminative brain regions for each classification task. However, the classification accuracy for MCIc vs. MCInc was relatively low, likely due to insufficient training samples and the heterogeneous nature of the MCI cohorts, warranting further investigation and optimization to improve performance and identify brain regions with stronger discriminability.

PET-based studies, using FDG-PET and amyloid PET, have also demonstrated the potential of AI in AD prediction. Kowsari et al. [38] applied a DL model to FDG-PET data from 338 subjects, achieving high accuracy in classifying AD (100%) and predicting MCI-to-AD conversion (82%). However, the study relied solely on PET and did not account for other potential biomarkers. While single-modality studies have shown promising results, they may not fully capture the complex heterogeneity of AD. Moreover, these studies often have limited sample sizes and may not be representative of the wider AD population [51].

### *Multi-modality studies: advantages, challenges, and notable contributions*

Multi-modality studies combine information from different neuroimaging techniques, as well as other biomarkers, to improve AD prediction performance and robustness. Integrating multi-modal data, such as neuroimaging, genetic, and clinical information, can significantly enhance the accuracy and reliability of predictive models in AD research by capturing different aspects of the disease process and leveraging complementary information [ 6, 52].

One key advantage of multi-modal data integration is the ability to capture different aspects of AD pathophysiology. Structural MRI reveals patterns of brain atrophy, while PET imaging provides information on brain metabolism and amyloid or tau accumulation. Combining these imaging modalities offers a more comprehensive picture of the neurodegeneration and pathological changes associated with AD [53]. Moreover, incorporating genetic data, such as APOE ε4 allele status, and clinical information, like cognitive test scores and neuropsychological assessments, can further enhance the predictive power of AI models [54].

For example, Zhang et al. [55] developed a multi-modal deep learning framework incorporating MRI, FDG-PET, and cognitive scores from 331 subjects, achieving high accuracy in classifying AD (93%) and predicting MCI-to-AD conversion (82%). Similarly, Paquerault et al. [56] combined MRI, FDG-PET, CSF biomarkers, and APOE genotype using a multi-kernel SVM approach on data from 402 subjects, achieving high accuracy in classifying AD (91%) and predicting MCI-to-AD conversion (82%). These studies demonstrate the superior performance of multi-modal approaches compared to single-modality models and traditional machine learning methods.

However, multi-modality studies also face challenges such as data heterogeneity, missing data, and the need for effective feature selection and fusion strategies [6, 52, 57]. Researchers are actively developing advanced AI techniques, such as multi-view learning, transfer learning, and graph convolutional networks, to address these challenges and optimize the integration of multi-modal data [58, 59].

In summary, while multi-modality studies have shown improved performance in AD prediction, they also face various challenges. Effective strategies for data integration and addressing associated challenges are crucial for realizing the full potential of multi-modal data in AD research. Future studies should focus on developing novel AI techniques to optimize the integration of diverse biomarkers and further enhance the accuracy and reliability of predictive models in AD.

### *Longitudinal studies and prediction of AD progression*

Longitudinal studies, which leverage repeated measurements of neuroimaging and clinical data, are crucial for understanding and predicting AD progression. Recent studies have applied AI techniques to longitudinal data for AD progression modeling. For example, Ghazi et al. [42] developed a deep learning approach using RNNs for longitudinal MRI data of 1145 subjects and achieved a mean absolute error of 1.7 years in predicting time to dementia onset for MCI patients. Bhagwat et al. [43] used a combination of deep learning and survival analysis to predict AD progression from baseline MRI and clinical data of 1010 subjects. However, neither study included other potentially relevant biomarkers, such as PET or CSF data. Longitudinal studies also face challenges such as missing data, irregular follow-up intervals, and the need for large, well-characterized datasets [60]. Moreover, the interpretation and clinical translation of these models can be complex, requiring close collaboration between AI researchers and clinicians.

Longitudinal studies, which are crucial for understanding and predicting AD progression, often face the challenge of missing data. Missing data can occur due to various reasons, such as participant dropout, missed visits, or data collection issues. When left unaddressed, missing data can lead to biased results and reduced statistical power, affecting the accuracy and reliability of predictive models [61].

Several strategies have been proposed to handle missing data in longitudinal AD studies. One common approach is complete case analysis, which only includes participants with complete data at all time points.

However, this method may result in a significant loss of information and biased estimates if the missing data is not random [62].

Another approach is imputation, which involves estimating the missing values based on the available data. Simple imputation methods, such as mean or last observation carried forward (LOCF), may introduce bias and underestimate the variability in the data [63]. More sophisticated imputation techniques, such as multiple imputation and machine learning-based methods, can provide more accurate estimates by considering the uncertainty associated with the missing values [64].

In the context of AI-based predictive models, missing data can affect model training and performance. Some machine learning algorithms, such as decision trees and random forests, can handle missing data directly by learning from the available data patterns [65]. Other methods may require imputation as a preprocessing step before model training.

Researchers have also developed AI-based techniques specifically designed to handle missing data in longitudinal AD studies. For example, Ghazi et al. [42] proposed a deep learning approach using RNNs that can effectively model disease progression in the presence of missing data. Their method outperformed traditional approaches, such as mean imputation and LOCF, in terms of predictive accuracy.

In summary, both single-modality and multi-modality studies have demonstrated the potential of AI in early AD prediction, with multi-modality approaches generally showing improved performance. Longitudinal studies have further highlighted the value of AI in modeling AD progression. However, these studies also face various challenges, including data harmonization, interpretability, and generalizability. While the evaluation metrics reported in AD prediction studies provide valuable insights into the performance of AI models, direct comparisons across studies should be made with caution. Considering the validation method, dataset characteristics, and choice of evaluation metrics is essential for a more nuanced interpretation of the results. Future studies should aim for more standardized reporting and validation practices to facilitate comparability and support the translation of AI models into clinical practice. Future research should focus on addressing these limitations, integrating diverse biomarkers, and developing explainable AI models to facilitate clinical translation (Table 2).

**Table 2** Overview of AI applications in early AD prediction using neuroimaging data

| Authors | Approach | N | Data Usage | Evaluation Method | Main Results |
|---|---|---|---|---|---|
| Basaia et al. (32) | Deep learning on structural MRI | 295 | 3D (static images) | tenfold cross-validation | High accuracy (98%) in classifying AD patients and healthy controls |
| Pan et al. (48) | Novel approach combining CNN and ensemble learning (EL) on structural MRI | 671 | 3D (static images) | fivefold cross-validation | Accuracies of 84% for AD vs. HC, 79% for MCIc vs. HC, and 62% for MCIc vs. MCInc; outperformed other methods (PCA + SVM, 3D-SENet) and identified discriminative brain regions |
| Ding et al. (33) | Deep learning on FDG-PET | 338 | 3D (static images) | Leave-one-out cross-validation | High accuracy in classifying AD (100%) and predicting MCI-to-AD conversion (82%) |
| Zhang et al. (50) | Multi-modal deep learning framework (MRI, FDG-PET, cognitive scores) | 331 | 3D (static images) | tenfold cross-validation | High accuracy in classifying AD (93%) and predicting MCI-to-AD conversion (82%) |
| Paquerault et al. (51) | Multi-kernel SVM (MRI, FDG-PET, CSF biomarkers, APOE genotype) | 402 | 3D (static images) | Nested cross-validation (outer tenfold, inner 5-fold) | High accuracy in classifying AD (91%) and predicting MCI-to-AD conversion (82%) |
| Ghazi et al. (40) | Deep learning (RNNs) on longitudinal MRI | 1145 | 4D (longitudinal 3D images) | Train-validation-test split (60%-20%) | Mean absolute error of 1.7 years in predicting time to dementia onset for MCI patients |
| Bhagwat et al. (41) | Deep learning and survival analysis on baseline MRI and clinical data | 1010 | 3D (static images) | fivefold cross-validation and external validation on ADNI dataset | Concordance index of 0.87 in predicting time to conversion from MCI to AD; external validation AUC of 0.81 on ADNI dataset |

# Challenges and future directions

Despite the significant progress in applying AI techniques to early AD prediction, several challenges remain to be addressed. These challenges span data standardization, model interpretability, generalizability, integration into clinical practice, and ethical considerations. Addressing these challenges is crucial for realizing the full potential of AI in AD research and clinical management.

## *Standardization of neuroimaging protocols and data preprocessing*

Standardizing neuroimaging protocols and data preprocessing is essential for ensuring the reproducibility and comparability of AI-based AD prediction models across different studies and populations. Initiatives such as the Alzheimer's Disease Neuroimaging Initiative (ADNI) [53] and the Australian Imaging, Biomarkers and Lifestyle (AIBL) study [68] have made significant efforts to harmonize neuroimaging data collection and preprocessing across multiple sites. However, further work is needed to establish and adopt standardized protocols for emerging neuroimaging techniques and to address data heterogeneity arising from differences in scanner types, acquisition parameters, and preprocessing pipelines [69]. Recommendations for standardization include developing and validating standardized protocols for emerging neuroimaging techniques, such as tau PET and functional connectivity MRI, to facilitate their integration into AI-based AD prediction models [70]. Encouraging the adoption of standardized data preprocessing pipelines, such as the ADNI pipeline [71], can minimize data heterogeneity and improve the comparability of AI models across studies. Promoting data sharing and collaborations among research groups to create large, diverse, and well-curated datasets for training and testing AI models is also essential.

## *Interpretability and explainability of AI models*

Improving the interpretability and explainability of AI models is crucial for building trust and facilitating their adoption in clinical practice. Many current AI models, particularly deep learning approaches, operate as "black boxes", making it difficult for clinicians and researchers to understand the basis for their predictions. Developing methods to visualize and interpret the features learned by AI models can help uncover the underlying biological mechanisms of AD and guide targeted interventions [72]. Recommendations for enhancing interpretability and explainability include encouraging the development and application of interpretable AI models, such as attention-based networks and graph convolutional networks, which can provide insights into the most informative regions and connections for AD prediction [73]. Integrating feature visualization techniques, such as class activation maps and saliency maps, can highlight the brain regions and patterns that contribute most to AI model predictions [74]. Developing user-friendly interfaces and visualization tools to present AI model interpretations in a clinically meaningful and actionable manner is also important.

## *Generalizability and robustness of AI algorithms*

Ensuring the generalizability and robustness of AI algorithms is essential for their successful translation into clinical practice. Many current AI models are trained and tested on relatively small and homogeneous datasets, which may not be representative of the diverse AD population encountered in real-world settings [51]. Validating AI models on large, independent, and diverse datasets is crucial for assessing their generalizability and robustness to variations in data quality, demographics, and disease subtypes [75]. Recommendations for improving generalizability and robustness include validating AI models on large, independent, and diverse datasets, such as the Global Alzheimer's Association Interactive Network (GAAIN) [76], to assess their generalizability across different populations and settings. Developing and applying data augmentation and transfer learning techniques can improve the robustness of AI models to variations in data quality and acquisition protocols. Investigating the performance of AI models across different AD subtypes and stages is also important to ensure their applicability to the full spectrum of the disease [77].

*Integration of AI tools into clinical practice*

Integrating AI tools into clinical practice requires addressing several challenges, including regulatory approval, data privacy and security, liability, and user acceptance. Developing clear guidelines for the development, validation, and deployment of AI tools in AD diagnosis and prognosis is essential for ensuring their safety, efficacy, and ethical use [78]. Collaborations between AI researchers, clinicians, and regulatory agencies are crucial for establishing standards and best practices for the clinical translation of AI tools [79]. Recommendations for clinical integration include developing clear regulatory guidelines and standards for the development, validation, and deployment of AI tools in AD diagnosis and prognosis, in collaboration with regulatory agencies such as the FDA and EMA [80]. Establishing data privacy and security protocols to ensure the confidentiality and protection of patient data used for AI model development and validation is also crucial. Engaging clinicians, patients, and caregivers in the design and development of AI tools can help ensure their usability, acceptability, and alignment with clinical needs and workflows.

*Ethical considerations and data privacy*

The use of AI in AD prediction raises important ethical considerations, including data privacy, informed consent, and transparency [81]. Ensuring that AI models are developed and applied in an ethical and responsible manner is crucial for maintaining public trust and maximizing their benefits for AD patients and society [82]. Addressing issues of bias and fairness in AI models is essential for preventing the exacerbation of health disparities and ensuring equitable access to AI-based interventions. Recommendations for addressing ethical considerations and data privacy include developing and adopting ethical guidelines and frameworks for the development and application of AI in AD research and clinical practice, in consultation with ethicists, patient advocates, and regulatory agencies. Ensuring transparency in the development and validation of AI models, including the disclosure of data sources, model architectures, and performance metrics, is also important. Implementing measures to detect and mitigate bias in AI models, such as using diverse and representative training datasets and conducting regular audits for fairness and non-discrimination, is crucial.

The limited availability of open access data poses a challenge for the development of robust and generalizable AI models in AD prediction. Large, diverse, and high-quality datasets are essential for training AI models that can accurately capture the heterogeneity of AD and its progression. However, many datasets used in AD research are proprietary or have restricted access, limiting the ability of researchers to share and collaborate on model development. Efforts to promote data sharing and the creation of open access databases, such as the Alzheimer's Disease Neuroimaging Initiative (ADNI) and the UK Biobank, are crucial for advancing AI-based AD prediction. Collaborative initiatives and public–private partnerships can help to pool resources and expertise, enabling the development of large, diverse, and representative datasets for AI research.

In conclusion, addressing the challenges of data standardization, model interpretability, generalizability, clinical integration, and ethical considerations is crucial for realizing the full potential of AI in early AD prediction. By implementing the recommendations outlined above, researchers and clinicians can work towards developing reliable, robust, and ethical AI tools that can improve the diagnosis, prognosis, and management of AD, ultimately benefiting patients and society.

**Funding** No funding was received for conducting this study.

**Data Availability** Not applicable.

**Ethics approval** Not applicable since no participants were recruited during the execution of this study.

**Human and animal rights** This article is based on previously conducted studies and does not contain any new studies with human participants or animals performed by any of the authors.

**Consent for publication** This manuscript has been approved for publication by all authors.

**Conflict of interest** The authors declare no competing interests.